\documentclass{article}

\usepackage{times}
\usepackage{graphicx} 
\usepackage{subfigure} 

\usepackage{natbib}

\usepackage{algorithm}
\usepackage{algorithmic}

\usepackage{amsmath}
\usepackage{amssymb}
\usepackage{amsthm}

\newtheorem{theorem}{Theorem}

\newtheorem{corollary}[theorem]{Corollary}

\theoremstyle{remark}

\newcommand{\FF}{{\mathcal{F}}}
\newcommand{\GG}{{\mathcal{G}}}
\newcommand{\LL}{{\mathcal{L}}}
\newcommand{\PP}{{\mathcal{P}}}

\newcommand{\DD}{{\mathcal{D}}}

\newcommand{\WW}{{\mathcal{W}}}
\newcommand{\XX}{{\mathcal{X}}}

\newcommand{\ZZ}{{\mathcal{Z}}}
\newcommand{\Real}{{\mathbb{R}}}

\DeclareMathOperator*{\expect}{\mathbb{E}}

\DeclareMathOperator*{\KL}{KL}

\graphicspath{{figures/}{../figures/}}

\usepackage{hyperref}

\usepackage[accepted]{icml2015}

\icmltitlerunning{Variational GSNs with Collaborative Shaping}

\begin{document} 

\twocolumn[
\icmltitle{Variational Generative Stochastic Networks with Collaborative Shaping}

\icmlauthor{Philip Bachman}{phil.bachman@gmail.com}
\icmladdress{McGill University, School of Computer Science}
\icmlauthor{Doina Precup}{dprecup@cs.mcgill.ca}
\icmladdress{McGill University, School of Computer Science}

\icmlkeywords{deep learning, manifold learning, neural networks, generative models, variational inference}

\vskip 0.3in
]

\begin{abstract} 

We develop an approach to training generative models based on unrolling a variational auto-encoder into a Markov chain, and shaping the chain's trajectories using a technique inspired by recent work in Approximate Bayesian computation. We show that the global minimizer of the resulting objective is achieved when the generative model reproduces the target distribution. To allow finer control over the behavior of the models, we add a regularization term inspired by techniques used for regularizing certain types of policy search in reinforcement learning. We present empirical results on the MNIST and TFD datasets which show that our approach offers state-of-the-art performance, both quantitatively and from a qualitative point of view.
\end{abstract}

\section{Introduction}
\label{sec:introduction}

Significant research effort has been directed towards developing models capable of effectively synthesizing samples from complicated distributions. We propose an approach to this problem whose goal is two-fold. We want to learn a distribution which is practically indistinguishable from the target distribution and we also want training, inference and sampling to be efficient. Our approach can be viewed as a class of Generative Stochastic Networks (GSNs)~\cite{Bengio2014}. We show that any model trained with the \emph{walkback} procedure~\cite{Bengio2013} is encompassed by our approach. Instead of using denoising auto-encoders, we leverage recent variational methods for deep, directed generative models, e.g.~\cite{Kingma2014a, Rezende2014, Mnih2014}, and build our approach starting from variational auto-encoders. By feeding the output of such an auto-encoder back into itself, we construct a Markov chain whose stationary distribution provably (in the non-parametric, infinite-data limit) converges to the target distribution. 

As an alternative to the walkback procedure for training GSNs, we propose an approach based on recent work in Approximate Bayesian Computation. We partner a generative model with a function approximator that estimates the log-density ratio between the model-generated distribution and the target distribution, in what can be seen as a \emph{collaborative} alternative to the \emph{adversarial} approaches in~\cite{Gutmann2014a, Gutmann2014b, Goodfellow2014}. We show that the global minimizer of the resulting objective (in the non-parametric, infinite-data limit) is achievable only when the model distribution matches the target distribution.

To control the model complexity, we introduce a regularization term close in spirit to reinforcement learning methods such as relative entropy policy search~\cite{Peters2010} and other approaches which depend on a notion of ``natural system dynamics'', e.g.~\cite{Todorov2009}. Specifically, we re-weigh the standard $\KL(\mbox{posterior} \, || \, \mbox{prior})$ term that appears in the variational free-energy. For a generative model $p(x) = \sum_z p(x | z) p(z)$, with latent variables $z \in \ZZ$, this approach provides a direct mechanism for trading complexity in $p(x)$ against the ability to exactly reproduce the training distribution whenever most of the complexity in $p(x)$ is captured by the latent variables. E.g.~if $p(x|z)$ is an isotropic Gaussian whose mean varies with $z$, but whose variance is the same for all $z$, restricting $\KL(p(z|x) \, || \, p(z))$ to be small for all $x$ forces $p(x)$ to be approximately an isotropic Gaussian. 

Our models permit efficient generation of independent samples, efficient generation of sequences of samples representing ``locally-coherent'' random walks along the data manifold, and efficient evaluation of a variational lower-bound on the log-likelihood assigned to arbitrary inputs. We show that our approach produces models which significantly outperform the GSNs in~\cite{Bengio2014} and the adversarial networks in~\cite{Goodfellow2014} in terms of test-set log-likelihood and qualitative behavior. 

\section{Background}
\label{sec:foundations}


This section provides a summary of prior work on denoising auto-encoders and Generative Stochastic Networks, which constitute the basis of our model.

\subsection{Generalized Denoising Auto-encoders}

In the Generalized Denoising Auto-encoder (DAE) framework~\cite{Bengio2013}, one trains a \emph{reconstruction distribution} $p_{\theta}(x | \tilde{x})$ to match the conditional distribution $\PP(x | \tilde{x})$ implicit in an infinite set of pairs $\{(x_1, \tilde{x}_1), ..., (x_n, \tilde{x}_n)\}$ generated by drawing each $x_i \in \XX$ from the \emph{target distribution} $\DD$ and then generating each $\tilde{x}_i \in \XX$ by applying some stochastic \emph{corruption process} $q_{\phi}(\tilde{x} | x)$ to $x_i$. Given $q_{\phi}$ and $p_{\theta}$, where $\theta$ and $\phi$ denote parameters of the two distributions, one can construct a Markov chain over $x \in \XX$ by iteratively sampling a point $x_t$ from $p_{\theta}(x_{t} | \tilde{x}_{t-1})$ and then sampling a point $\tilde{x}_{t}$ from $q_{\phi}(\tilde{x}_{t} | x_{t})$. The chain is initialized at $t = 0$ by sampling $x_{0}$ directly from $\DD$ and its transition operator $T_{\theta}(x_{t} | x_{t-1})$  can be computed by marginalizing over $\tilde{x}_{t-1}$.

Given a few small assumptions on the forms of $q_{\phi}$ and $p_{\theta}$, and the larger assumption that $p_{\theta}(x | \tilde{x})$ provides a consistent estimator of $\PP(x | \tilde{x})$ as the number of training samples $x \sim \DD$ goes to infinity, it was shown in \cite{Bengio2013} that the Markov chain constructed from the iterative process described above will be ergodic and have a stationary distribution $\pi_{\theta}$ which matches $\DD$ (i.e.~$\forall x, \pi_{\theta}(x) = \DD(x)$).

\begin{figure*}
\begin{center}
  \subfigure[]{\includegraphics[scale=0.32]{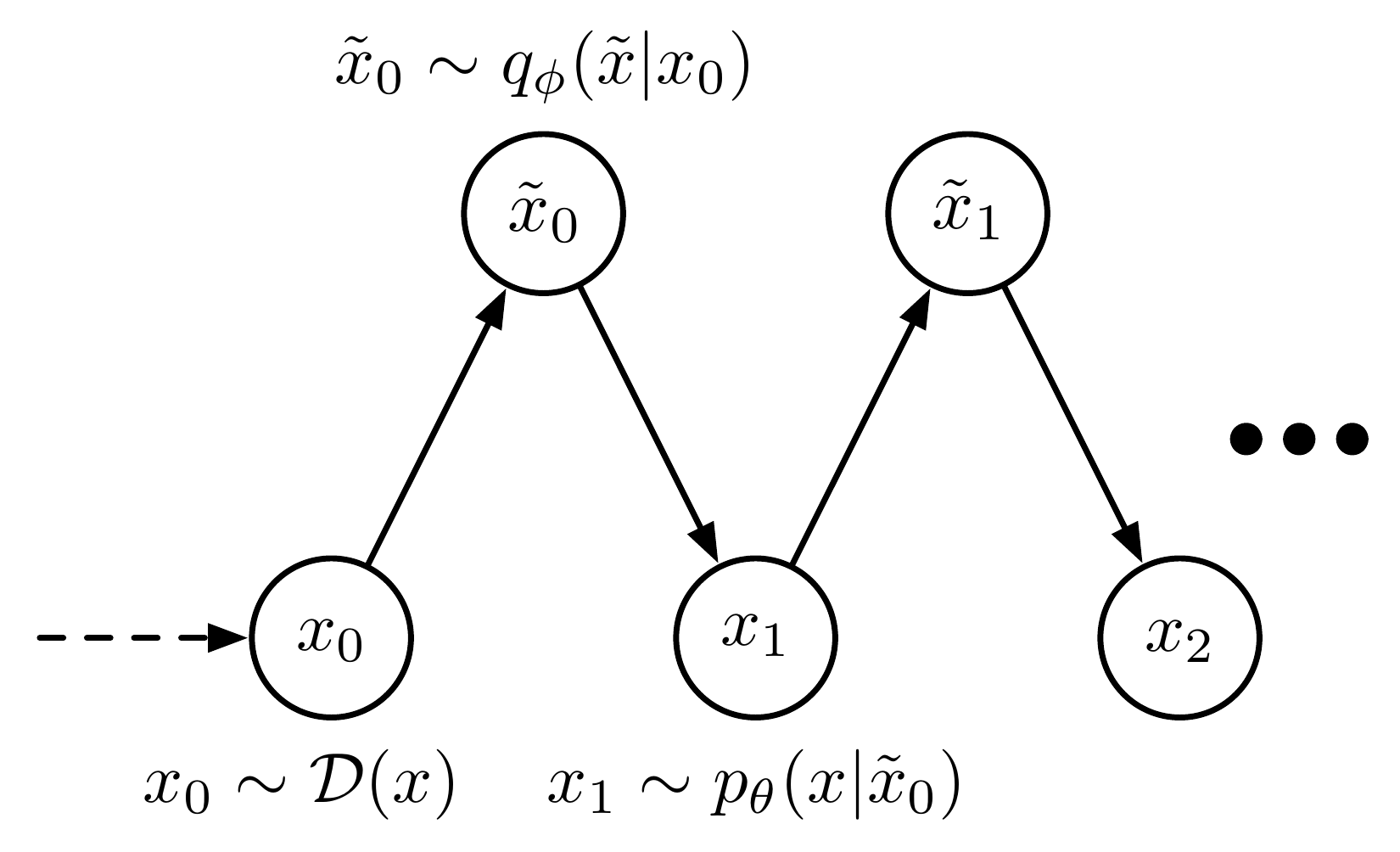}}
  \subfigure[]{\includegraphics[scale=0.32]{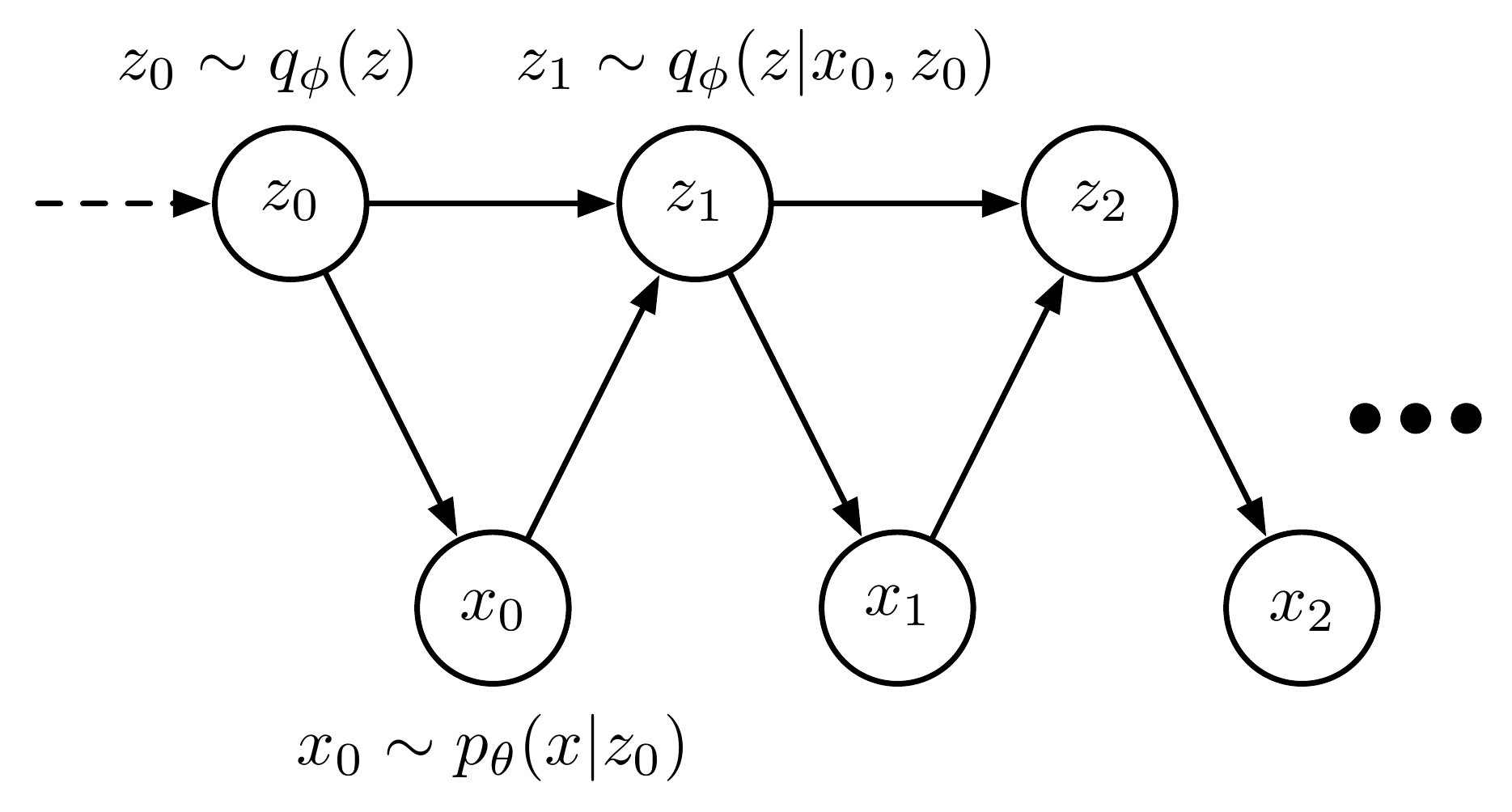}}
  \subfigure[]{\includegraphics[scale=0.32]{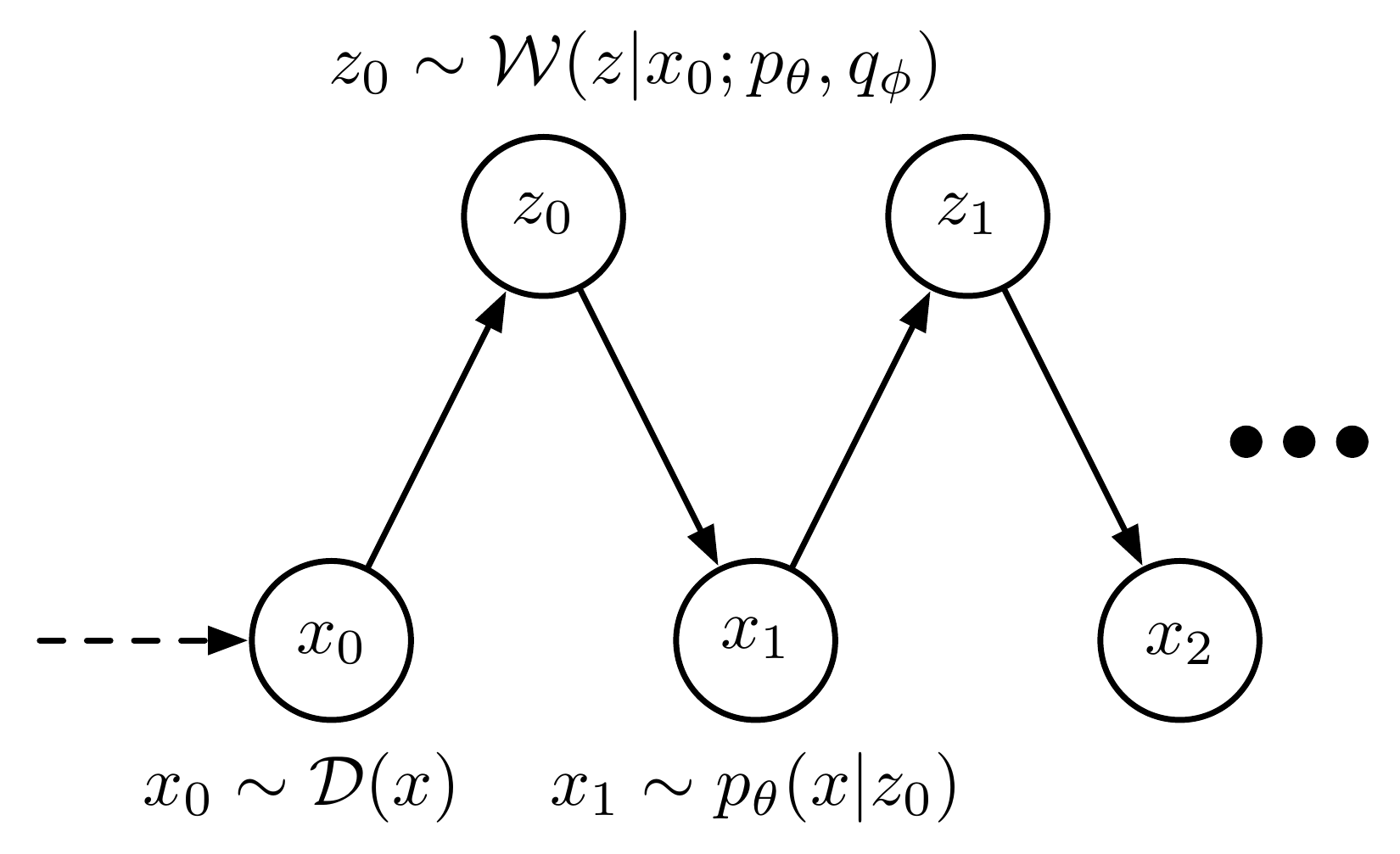}}
  \caption{(a) shows how to construct the Markov chain associated with a Generalized DAE with reconstruction distribution $p_{\theta}$, corruption process $q_{\phi}$, and target distribution $\DD$. (b) shows how to construct the Markov chain associated with a GSN with reconstruction distribution $p_{\theta}$ and corruption process $q_{\phi}$. We overload notation and define $q_{\phi}(z)$ to be the distribution matching the stationary distribution of the GSN chain over $z$. (c) shows how to construct the Markov chain associated with a Simple GSN that uses a reconstruction distribution $p_{\theta}$ and a process $\WW(z | x; q_{\phi}, p_{\theta})$ formed by wrapping $p_{\theta}$ and a corruption process $q_{\phi}$ through the walkback procedure (see text for details).}
\end{center}
\label{fig:gdae_and_gsn_graphs}
\end{figure*}

All of the discussion in~\cite{Bengio2013} assumed that both $x_i$ and $\tilde{x}_i$ for each training pair $(x_i, \tilde{x}_i)$ were from the same space $\XX$, although this was not required for their proofs. The Generative Stochastic Network (GSN) framework~\cite{Bengio2014} thus made the jump of assuming a corruption process $q_{\phi}(z_{t} | x_{t-1}, z_{t-1})$. This extends the Generalized DAE framework by introducing a \emph{latent} space $\ZZ \neq \XX$, and by allowing the current latent state $z_t$ to depend on the previous latent state $z_{t-1}$ (in addition to its dependence on the previous \emph{observable} state $x_{t-1}$). Figures~2 (a) and (b) illustrate the graphical models corresponding to Generalized DAEs and GSNs.

\subsection{Training with Walkback}

\begin{algorithm}[tb]
   \caption{Walkback for a General GSN}
   \label{alg:gsn_walkback}
\begin{algorithmic}
   \STATE {\bfseries Input:} data sample $x$, corruptor $q_{\phi}$, reconstructor $p_{\theta}$
   \STATE Initialize an empty training pair list $\PP_{xz} = \{ \, \}$
   \STATE Set $\hat{z}$ to some initial vector in $\ZZ$.
   \FOR{$i=1$ {\bfseries to} $k_{burn-in}$}
   \STATE Sample $\check{z}$ from $q_{\phi}(z | x, \hat{z})$ then set $\hat{z}$ to $\check{z}$.
   \ENDFOR
   \STATE Set $\hat{x}$ to $x$.
   \FOR{$i=1$ {\bfseries to} $k_{roll-out}$}
   \STATE Sample $\check{z}$ from $q_{\phi}(z | \hat{x}, \hat{z})$ then set $\hat{z}$ to $\check{z}$.
   \STATE Sample $\check{x}$ from $p_{\theta}(x | \hat{z})$ then set $\hat{x}$ to $\check{x}$.
   \STATE Add pair $(x, \hat{z})$ to $\PP_{xz}$.
   \ENDFOR
   \STATE {\bfseries Return: $\PP_{xz}$}.
\end{algorithmic}
\end{algorithm}

A method called \emph{walkback} training was proposed for Generalized DAEs in~\cite{Bengio2013} and used again for GSNs in~\cite{Bengio2014}. The motivation for walkback training was to mitigate difficulties encountered in practical, finite-data settings, where many values for the latent variables $z \in \ZZ$ that were rarely (if ever) visited during training would appear when sampling from the resulting Markov chain. The difficulties stem primarily from a desire to make the corruption process $q_{\phi}$ induce a conditional distribution $\PP(x|z)$ which is roughly unimodal over $x$ given any particular $z$ (because this makes it easier to model with $p_{\theta}(x | z)$), which fights against the possibility that $\DD$ contains multiple well-separated modes (which would necessitate a relatively non-local corruption process, able to ``carve out'' reliable paths between these modes during training).

The walkback procedure can be interpreted as a ``wrapper'' function which takes the corruption process $q_{\phi}$ and the reconstruction distribution $p_{\theta}$, and then samples from a  process $\WW(z | x; q_{\phi}, p_{\theta})$ which procedurally generates a distribution over $z \in \ZZ$ given any $x \in \XX$, as shown in Alg.~\ref{alg:gsn_walkback}. For example, in the original GSN paper \cite{Bengio2014}, the reconstruction distribution $p_{\theta}(x | z)$ for a GSN which emulates Gibbs sampling in a Deep Boltzmann Machine was trained on pairs $(x, z)$ sampled from the walkback process described in Alg~\ref{alg:gsn_walkback}. The returned set of pairs  $\PP_{xz}$ can be viewed as containing data $(x, \hat{z}) \sim \WW(z | x; q_{\phi}, p_{\theta})$, where $\WW$ is specified procedurally rather than directly. The reconstruction distribution $p_{\theta}(x | z)$ is then trained to approximate the conditional $\PP_{xz}(x | z)$ implicit in the pairs generated via Alg.~\ref{alg:gsn_walkback}.


\section{Simple Generative Stochastic Networks}
\label{sec:sgsn}

We define a ``Simple GSN'' as any GSN in which the corruption process  renders $z_t$ independent of $z_{t-1}$ given $x_{t-1}$. Simple GSNs thus represent the minimal, direct extension of Generalized DAEs to corruption processes that may produce outputs in a different space from their inputs. Fig.~1(c) shows the structure of a simple GSN based on iteratively sampling from a walkback  process $\WW(z | x; q_{\phi}, p_{\theta})$ and a reconstruction distribution $p_{\theta}(x | z)$.

The Simple GSN model is in fact quite general, and covers all GSNs trained with a walkback procedure. We now give versions of the theorems from~\cite{Bengio2013} modified for Simple GSNs, which show that training with enough data and with sufficiently powerful function approximators $p_{\theta}/q_{\phi}$ produces a Markov chain whose asymptotic distribution exists and matches the target distribution.

\begin{theorem}
If $p_{\theta}(x | z)$ is a consistent estimator of the true conditional distribution $\PP(x | z)$ and the transition operator $T_{\theta}(x_{t+1} | x_{t})$ that samples $z_{t}$ from $q_{\phi}(z_{t} | x_{t})$ and $x_{t+1}$ from $p_{\theta}(x_{t+1} | z_{t})$ defines an ergodic Markov chain, then as the number of examples used to train $p_{\theta}(x | z)$ goes to infinity (i.e.~as $p_{\theta}(x | z)$ converges to $\PP(x|z)$), the asymptotic distribution of the Markov chain defined by $T_{\theta}$ converges to the target distribution $\DD$.
\end{theorem}

The proof is a direct translation of the proof for Theorem 1 in \cite{Bengio2013}, but with $z$s replacing $\tilde{x}$s. Briefly, drawing an initial $x$ from $\DD$ and then sampling alternately from $q_{\phi}(z | x)$ and $\PP(x | z)$ is equivalent to sampling from a Gibbs chain for the joint distribution over $(x_i, z_i)$ generated by repeatedly sampling $x_i \sim \DD$ and  $z_i \sim q_{\phi}(z | x_i)$. Ergodicity guarantees the existence of an asymptotic distribution 
for the Markov chain. Since $p_{\theta}(x | z)$ converges to the true $\PP(x | z)$, the asymptotic distribution of the chain converges to the marginal distribution of $x$ in the Gibbs chain, which is just $\DD(x)$.

\begin{corollary}
Let $\XX$ be a set in which every pair of points is connected by a finite-length path contained in $\XX$.  Suppose that for each $x \in \XX$ there exists a ``shell" set ${\cal S}_x \subseteq \XX$  such that  all paths between $x$ and any point in $\XX \setminus {\cal S}_x$ pass through some point in ${\cal S}_x$ whose shortest path to $x$ has length $> 0$.  Suppose that  $\forall x^{\prime} \in {\cal S}_x \cup \{x\} ,\exists z_{xx^{\prime}}$ such that $q_{\phi}(z_{xx^{\prime}} | x) > 0$ and $p_{\theta}(x^{\prime} | z_{xx^{\prime}}) > 0$.  Then, the Markov chain with transition operator $T_{\theta}(x_{t+1} | x_{t}) = \sum_{z} p_{\theta}(x_{t+1} | z) q_{\phi}(z | x_{t})$ is ergodic.
\end{corollary}

\begin{proof}
The chain is aperiodic because the assumptions imply that $\forall x, \exists z_{xx}$ such that $q_{\phi}(z_{xx} | x) > 0$ and $p_{\theta}(x | z_{xx}) > 0$, so $T_{\theta}(x_{t+1}=x | x_{t}=x) > 0$. To show that the chain is irreducible, note that by assumption, $\forall x'\in {\cal S}_x, T(x_{t+1}=x'|x_t=x)>0$.  For any $x'\not\in {\cal S}_x$, consider the shortest path from $x$ to $x'$ and note that $\exists y \in {\cal S}_x$ on this path such that the shortest path between $x$ and $y$ is $>0$ and $T_{\theta}(x_{t+1}=y | x_{t}=x)>0$. Hence, $T(x_{t+1}=x'|x_t=x)>0$, as the path $x \rightarrow x^{\prime}$ can be decomposed into a finite sequence of finite-length segments, each with non-zero transition probability.
Because the chain is over a finite state space it is also  positive recurrent. As the chain  is aperiodic, irreducible, and positive recurrent, it is also ergodic. 
\end{proof}

The restricted dependency structure of Simple GSNs lets us avoid some complications faced by the proofs in~\cite{Bengio2014}. Our proof of Corollary 2 also avoids reliance on an $\epsilon$ ball, which does not work correctly in discrete or discontinuous spaces, in which paths starting at $x$ with length $> \epsilon$ may contain no ``segments'' overlapping with the set of all paths starting at $x$ with length $\leq \epsilon$. 

Training $p_{\theta}(x | z)$ for any GSN using samples generated by walkback as described in Algorithm~\ref{alg:gsn_walkback} corresponds to training a Simple GSN built around the reconstruction distribution $p_{\theta}$ and corruption process $\WW(z | x; q_{\phi}, p_{\theta})$. The key observation here is that the samples in $\PP_{xz}$ generated by Algorithm~\ref{alg:gsn_walkback} are obtained by relating \emph{multiple} sampled latent variables $\hat{z}$ back to the \emph{single} observable variable $x$ given as input. In order to train $p_{\theta}$ to be consistent with the joint distribution over $(x, z)$ pairs generated by the Markov chain constructed from $p_{\theta}$ and $q_{\phi}(z_{t} | x_{t-1}, z_{t-1})$, it would actually be necessary to train $p_{\theta}$ on pairs $(x_{t-1}, z_{t})$ generated by explicitly unrolling the chain. In Sec.~\ref{sec:collaborative} we present a mechanism based on Approximate Bayesian Computation that allows training $p_{\theta}$ directly on the pairs $(x_{t-1}, z_{t})$ generated by unrolling a GSN's Markov chain and applying backpropagation through time (BPTT).

Walkback can be viewed as an effective way to construct a more dispersed distribution over the latent space $\ZZ$ than would be provided by the original corruption process $q_{\phi}$. Though not explicitly stated in the existing work on GSNs, it seems that balancing between maximizing dispersion of the corruption process and the ease of modeling the reconstruction distribution $p_{\theta}$ plays a role for GSNs analogous to balancing between minimizing the KL divergence $\KL(q_{\phi}(z | x) || p(z))$ and maximizing the expected conditional log-likelihood $\expect_{z \sim q_{\phi}(z | x)} \log p_{\theta}(x|z)$ when training a generative model $p_{\theta}(x)$ with variational methods, or balancing between following the ``natural dynamics" of the system and optimizing reward in policy search~\cite{Peters2010, Todorov2009}. The next section expands on this relation.

\section{Variational Simple GSNs}
\label{sec:variational}

We now develop a Simple GSN which can efficiently generate ``locally-coherent'' random walks along the manifold described by the target distribution $\DD$, efficiently generate independent (approximate) samples from the target distribution $\DD$, and efficiently evaluate a lower-bound on the log-likelihood assigned by the model to arbitrary inputs. We do this by replacing the denoising auto-encoders in existing examples of Generalized DAEs and GSNs with variational auto-encoders~\cite{Kingma2014a, Rezende2014}, while reinterpreting the two competing terms in the variational free-energy $\FF$ (see Eq.~\ref{eq:var_free_energy_1}) as representing an explicit trade-off between the dispersion of $q_{\phi}(z|x)$ and the ease of modeling $p_{\theta}(x | z)$.

Suppose that, in addition to  $p_{\theta}(x | z)$ and  $q_{\phi}(z | x)$, we also have access to  a distribution $p_{\ast}(z)$ over $\ZZ$ (which could be learned or fixed a priori). Given these three distributions, we can define the \emph{derived distribution} $p_{\theta}(x; p_{\ast})$ such that $p_{\theta}(x; p_{\ast}) = \sum_{z} p_{\theta}(x | z) p_{\ast}(z)$, and the variational free-energy $\FF(x; q_{\phi}, p_{\theta}, p_{\ast})$, which provides an upper-bound on the negative log-likelihood of $x \in \XX$ under  $p_{\theta}(x; p_{\ast})$:
\begin{align}
& \FF(x; q_{\phi}, p_{\theta}, p_{\ast}) =\\
&= -\sum_{z} [q_{\phi}(z | x) \log p_{\theta}(x | z)] \label{eq:var_free_energy_1} 
+ \KL(q_{\phi}(z | x) || p_{\ast}(z)) \nonumber \\
&\geq -\log p_{\theta}(x; p_{\ast})
\end{align}
A step-by-step derivation is provided in the Appendix.

Given $\FF(x; q_{\phi}, p_{\theta}, p_{\ast})$, we can maximize a lower-bound on the expected log-likelihood of samples $x \sim \DD$ under the model $p_{\theta}(x; p_{\ast})$ by minimizing:
\begin{eqnarray}
&\expect_{x \sim \DD}& \FF(x; q_{\phi}, p_{\theta}, p_{\ast}) = \label{eq:vae_objective} \\
&\expect_{x \sim \DD}& \left[ \expect_{z \sim q_{\phi}(z | x)} [-\log p_{\theta}(x | z)] + \KL(q_{\phi}(z | x) || p_{\ast}(z)) \right] \nonumber
\end{eqnarray}
It is useful to compare this to the objective for Generalized DAEs, i.e.~Eq.~4 in~\cite{Bengio2013}:
\begin{equation}
\LL(\theta) = \expect_{x \sim \DD, \tilde{x} \sim q_{\phi}(\tilde{z} | z)} \left[ - \log p_{\theta}(x | \tilde{x}) + \lambda \Omega(\theta, x, \tilde{x}) \right] \label{eq:generalized_dae_objective}
\end{equation}
in which $\Omega(\theta, x, \tilde{x})$ is a regularization term for controlling the capacity of $p_{\theta}$ when the number of available samples $x \sim \DD$ is finite. The basic training process for both Eq.~\ref{eq:vae_objective} and \ref{eq:generalized_dae_objective} can be described as follows: draw a sample $x \sim \DD$, apply a random corruption to get $z/\tilde{x} \sim q_{\phi}(\cdot | x)$, then adjust the parameters to reduce $-\log p_{\theta}(x | \cdot)$ and an added regularization term. Training with walkback simply changes the $\tilde{x} \sim q_{\phi}(\tilde{z} | z)$ in Eq.~\ref{eq:generalized_dae_objective} to $\tilde{x} \sim \WW(\tilde{x} | x; p_{\theta}, q_{\phi})$.

We consider the objective in Eq.~\ref{eq:vae_objective} from a GSN perspective and treat it as comprising two terms: a reconstruction error $-\log p_{\theta}(x | \cdot)$ and a dispersion maximization term $\KL(q_{\phi}(\cdot|x) \, || \, p_{\ast})$. Based on a desire to keep $q_{\phi}$ well-dispersed for all $x \in \XX$, and to keep the degree of the dispersion relatively consistent across $x \in \XX$, we replace the basic $\KL$ term in Eq.~\ref{eq:vae_objective} with the following:
\begin{equation}
\lambda \left( \KL(q_{\phi}(\cdot|x) || p_{\ast}) + \gamma ([\KL(q_{\phi}(\cdot|x) || p_{\ast}) - \bar{K}]_{+})^2 \right), 
\label{eq:dispersion_reg}
\end{equation}
in which $\bar{K}=\expect_{x\sim\XX} \KL(q_{\phi}(\cdot|x) || p_{\ast})$ and $[\cdot]_{+}$ indicates positive rectification, i.e.~clamping all negative values to $0$. This penalizes both the magnitude and variance of the per-example $\KL$, while avoiding any pressure to increase $\KL$. We make this modification to allow tighter control over the trade-off between reconstruction fidelity and dispersion of the corruption process. Although re-weighting the $\KL$ term in Eq.~\ref{eq:vae_objective} may seem odd to those already familiar with variational methods, we emphasize that the free-energy $\FF(x; q_{\phi}, p_{\theta}, p_{\ast})$ from Eq.~\ref{eq:var_free_energy_1} still provides a valid upper-bound on $-\log p_{\theta}(x; p_{\ast})$. The appendix provides further discussion of this variational free-energy.

\section{Collaborative Generative Networks}
\label{sec:collaborative}

In this section we take a step back and present a general method for shaping the distribution $\GG$ produced by a generative model $g_{\theta}$ to be \emph{practically} indistinguishable from a target distribution $\DD$. In Section~\ref{sec:perambulate} we combine this method with variational Simple GSNs to directly train unrolled Markov chains. The general approach of estimating the parameters of $g_{\theta}$ to minimize some computable measure of dissimilarity between $\GG$ and $\DD$ is called Approximate Bayesian Computation. Examples of Approximate Bayesian Computation include spectral methods based on the ``method of moments'', which learn the parameters of $g_{\theta}$ so as to match some of the statistical moments of $\GG$ to the corresponding moments observed in an empirical sample from $\DD$, and more recent approaches based on minimizing the ability of some classifier to distinguish between $\GG$ and $\DD$~\cite{Gutmann2014a, Gutmann2014b, Goodfellow2014}. Motivated by the recent empirical success of this latter approach in training deep generative models~\cite{Goodfellow2014}, we develop a related approach which offers improved stability and a simpler proof of correctness.

We define an objective for jointly optimizing a \emph{generator} function $g_{\theta}$ and a \emph{guide} function $f_{\psi}$ which shapes the distribution $\GG$ produced by $g_{\theta}$ to match a target distribution $\DD$. Our method can be interpreted as a collaboration between $g_{\theta}$ and $f_{\psi}$, in contrast with the adversarial approach presented in~\cite{Goodfellow2014}. It is based on optimizing a term which encourages $f_{\psi}$ to approximate the log-density-ratio $\log \frac{\DD(x)}{\GG(x)}$ for $x \in \XX$ while also using $f_{\psi}$ as a guidance signal for redistributing the mass emitted by $g_{\theta}$. Our objective comprises two parts: one optimized by $f_{\psi}$ and the other optimized by $g_{\theta}$. We show that $g_{\theta}$ and $f_{\psi}$ can simultaneously minimize their respective objectives if and only if $\forall x$, $\GG(x) = \DD(x)$ and $f_{\psi}(x) = \log \frac{\DD(x)}{\GG(x)} = 0$.

The objective $\LL_{f}$ for $f_{\psi}$ is the basic logistic regression loss for a binary classifier which assumes equal prior probability for the positive class $\DD$ and the negative class $\GG$, i.e.:
\begin{equation}
\LL_{f} = \expect_{x \sim \DD} \left[ b(f_{\psi}(x)) \right] + \expect_{x \sim \GG} \left[ b(-f_{\psi}(x)) \right] \label{eq:f_objective}
\end{equation}
where $b(f) = \log(\exp(-f) + 1)$ is the binomial deviance loss. The objective $\LL_{g}$ for $g_{\theta}$ is based on, e.g., a one-sided absolute value loss:
\begin{equation}
\LL_{g} = \expect_{x \sim \GG} \left[ |f_{\psi}(x)| \cdot \mathbb{I}[f_{\psi}(x) < 0] \right] \label{eq:g_objective}
\end{equation}
in which $\mathbb{I}[ \cdot ]$ is a binary indicator function.

\begin{theorem}
The objectives $\LL_{f}$ and $\LL_{g}$ are simultaneously optimized with respect to $f_{\psi}$ and $g_{\theta}$ if and only if $\forall x \in \XX, \GG(x) = \DD(x)$ and $f_{\psi}(x) = \log \frac{\DD(x)}{\GG(x)} = 0$.
\end{theorem}

\begin{proof}
By definition of $\LL_{f}$, it is minimized w.r.t.~$f_{\psi}$ if and only if $\forall x, \; f_{\psi}(x) = \log \frac{\DD(x)}{\GG(x)}$ \cite{Hastie2008}. If $\LL_{f}$ is minimized w.r.t.~$f_{\psi}$ then we know furthermore  that either $\forall x, \; f_{\psi}(x) = \log \frac{\DD(x)}{\GG(x)} = 0$ or $\exists x$ s.t.~$f_{\psi}(x) = \log \frac{\DD(x)}{\GG(x)} > 0$ and $\exists x^{\prime}$ s.t.~$f_{\psi}(x^{\prime}) = \log \frac{\DD(x^{\prime})}{\GG(x^{\prime})} < 0$. The former situation results in $\LL_{g} = 0$. The latter situation results in $\LL_{g} > 0$, because $|f_{\psi}(x)| \cdot \mathbb{I}[f_{\psi}(x) < 0] \geq 0$ with equality only when $f_{\psi}(x) \geq 0$. Thus, whenever $\LL_{f}$ is optimized w.r.t.~$f_{\psi}$, $\LL_{g}$ can obtain its minimum possible value of $0$ if and only if $\forall x, \; \DD(x) = \GG(x)$.
\end{proof}

Note that this proof works for any $\LL_{g}$ which involves an expectation (w.r.t.~$x \sim \GG$) over a quantity which is everywhere non-negative and equal to $0$ if and only if $f_{\psi}(x) \geq 0$. We leave a detailed investigation of the relative merits of the various possible $\LL_{g}$ for future work. 

The key characteristic that distinguishes our objective from~ \cite{Goodfellow2014} is that, given a fixed guide function $f_{\psi}$, our objective for the generator $g_{\theta}$ pushes the mass in over-dense regions of $\GG$ towards zero-contours of $f_{\theta}$ while leaving the mass in under-dense regions unmoved. In contrast, the adversarial objective in~\cite{Goodfellow2014} drives all mass emitted by $g_{\theta}$ towards local maxima of $f_{\psi}$. These local maxima will typically correspond to points in $\XX$, while the zero-contours sought by our objective will correspond to regions in $\XX$. We can also incorporate additional terms in $\LL_{g}$, under the restriction that the added terms are minimized when $\forall x, \; \DD(x) = \GG(x)$. E.g.~moment matching terms for matching the mean and covariance of $x \sim \GG$ with those of $x \sim \DD$ can be included to help avoid the occasional empirical ``collapses'' of $\GG$ that were described in~\cite{Goodfellow2014}.


\section{Generating Random Walks on Manifolds}
\label{sec:perambulate}

\begin{figure}
\begin{center}
  \includegraphics[scale=0.33]{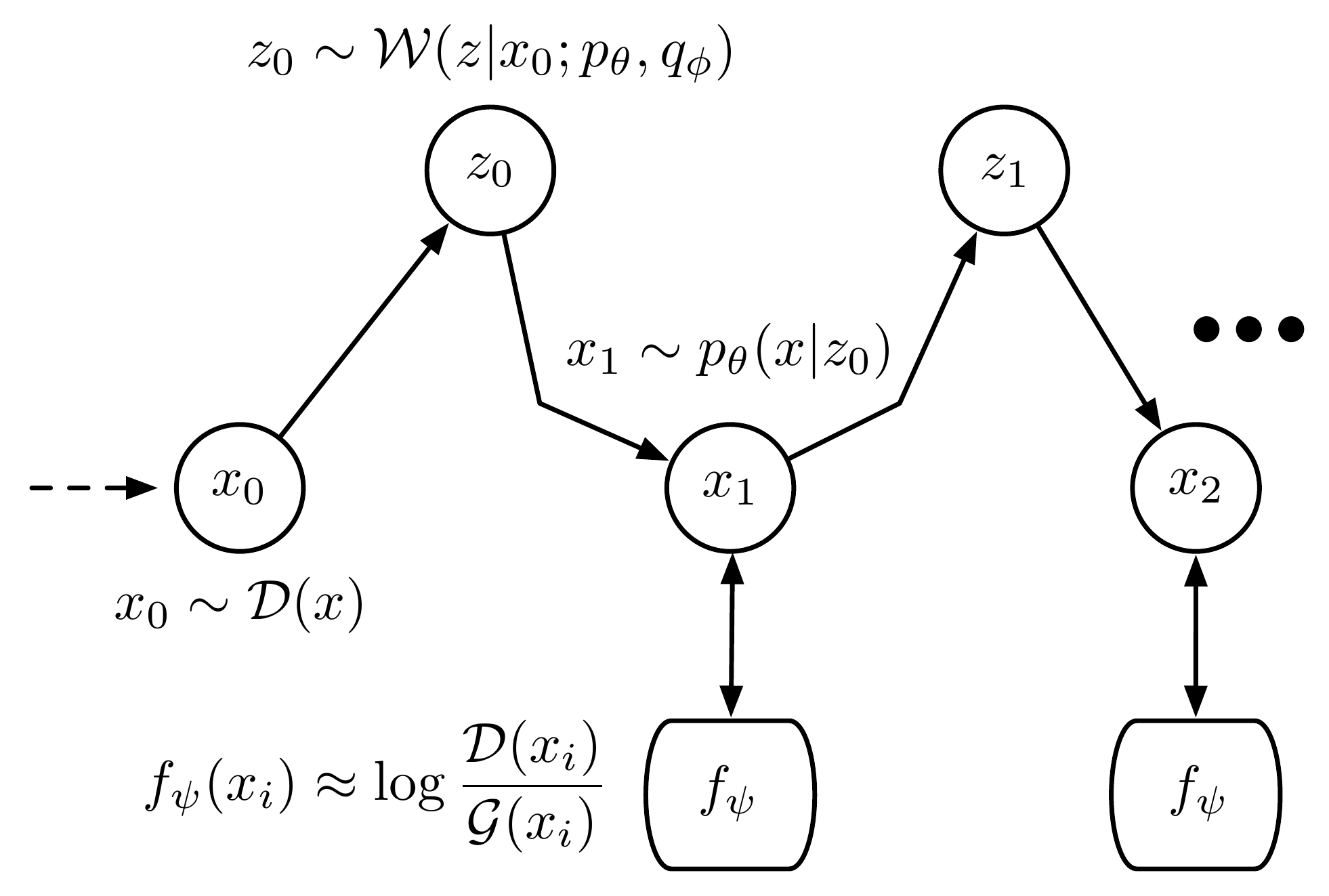}
  \caption{The VCG Loop: a self-looped variational auto-encoder whose asymptotic distribution is shaped by a guide function $f_{\psi}$ using the collaborative mechanism described in Sec.~\ref{sec:collaborative}.}
\label{fig:vcg_loop}
\end{center}
\end{figure}

We now combine the variational Simple GSNs from Sec.~\ref{sec:variational} with the collaborative mechanism from Sec.~\ref{sec:collaborative}. Our goal is to directly train the Markov chain constructed by unrolling the variational Simple GSN to produce locally-contiguous walks along the manifold of the target distribution $\DD$, and to have the asymptotic distribution of the chain approximate $\DD$.

The collaborative mechanism described in Sec.~\ref{sec:collaborative} pairs a generator $g_{\theta}$ with a guide function $f_{\psi}$. For the generator, we propose using a variational Simple GSN, which we unroll into a Markov chain by initializing with a sample $x_0 \sim \DD$ and then repeatedly generating $\{x_1, ..., x_{t}, ..., x_{n} \}$ by sampling  $z_{t} \sim q_{\phi}(z | x_{t})$  $x_{t+1} \sim p_{\theta}(x | z_{t})$. In other words, we self-loop a variational auto-encoder by piping its output back into its input.  The unrolled joint system is depicted in Fig.~2. In practice, we implement $p_{\theta}$, $q_{\phi}$  and the guide function $f_{\psi}$ using neural networks, whose specific architectures are described further in Sec.~\ref{sec:experiments}.

\section{Experiments}
\label{sec:experiments}

\begin{figure}
\begin{center}
  \subfigure[]{\includegraphics[scale=0.4]{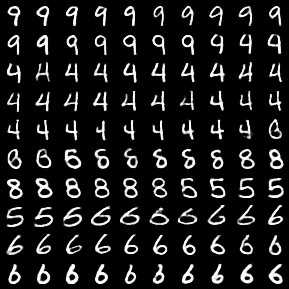}}
  \subfigure[]{\includegraphics[scale=0.4]{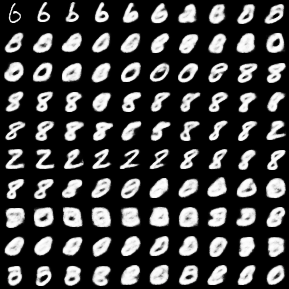}\vspace*{-5mm}}\vspace*{-5mm}
  \caption{Comparing the chains generated by models learned with and without collaboratively-guided unrolling. The samples in (a) were generated by a corrupt-reconstruct pair $q_{\phi}/p_{\theta}$ trained for 100k updates as a variational auto-encoder (VAE), and then 100k updates as a 6-step unrolled chain guided by a function $f_{\psi}$ as described in Sec.~\ref{sec:collaborative}. The samples in (b) were generated by a model with the same architecture and hyperparameters as in (a), but with 200k updates of VAE training. These chains are downsampled 5x.}
\end{center}
\label{fig:mnist_walkout}
\vspace{-0.5cm}
\end{figure}

We now present tests examining the behavior of  our models on the MNIST and TFD datasets. We chose these datasets to allow direct comparison with~\cite{Bengio2014} and \cite{Goodfellow2014}. 

Our first tests with MNIST data examined the benefits of training using the unrolled collaborative mechanism in Fig.~2. We represented $q_{\phi}$ and $p_{\theta}$ using neural networks with two hidden layers of 1500 rectified-linear units and we set the latent space $\ZZ$ to $\Real^{64}$. We used a Gaussian with identity covariance for the prior distribution $p_{\ast}(x)$. The output layer of $q_{\phi}$ produced two vectors in $\Real^{64}$, one representing the mean of a Gaussian distribution over $\ZZ$ and the other representing the element-wise log-variances of the distribution. Given this $q_{\phi}/p_{\ast}$, $\KL(q_{\phi}(z|x) || p_{\ast})$ was easy to compute analytically, and its gradients with respect to $\phi$ were readily available. We interpreted $p_{\theta}(x | z)$ as a factored Bernoulli distribution, with Rao-Blackwellisation over the possible binarizations of each $x$. I.e., the output layer of $p_{\theta}$ produced a vector in $\Real^{784}$, which was then passed through a sigmoid to get $\hat{x}$.  We minimized $-\log p_{\theta}(x | z) = -\mbox{sum} (x \odot \log \hat{x} + (1-x) \odot \log (1-\hat{x}))$, where $\odot$ denotes element-wise multiplication and we sum the vector entries. The gradients of $-\log p_{\theta}(x | z)$ w.r.t.~$\theta$ were directly available, and we backpropped through sampling $z \sim q_{\phi}(z | x)$ to get gradients w.r.t.~$\phi$ using the techniques in \cite{Kingma2014a, Rezende2014}. 

We used a Maxout network~\cite{Goodfellow2013} with two hidden layers of 1200 units in 300 groups of 4 for the guide function $f_{\psi}$. For $\LL_g$ in Eq.~\ref{eq:g_objective}, we used a half-rectified elastic-net \cite{Zou2005}, with the linear and quadratic parts weighted equally. We unrolled our chains for 6 steps. The distributions $\DD$ and $\GG$ for training $f_{\psi}$ according to Eq.~\ref{eq:f_objective} were given by the MNIST training set and the $x_i$ emitted by the unrolled chains. We passed gradients through the unrolled computation graph via BPTT.

We performed model updates using gradients computed from mini-batches of 100 distinct examples from the training set, each of which was passed through the model for 5 samples from $q_{\phi}(z|x)$. We trained using plain SGD. We pre-trained $p_{\theta}$ and $q_{\phi}$ as a variational auto-encoder (VAE) for 100k updates by running the model in Fig.~1(c) out to $x_1$. After 100k VAE updates we ``forked'' the model into a ``multi-step guided'' model and a ``one-step VAE'' model and performed a further 100k updates to each of the now-independent models. We implemented our models in Python using the THEANO library~\cite{Bergstra2010}. Code implementing the models described in this paper is available online at: \texttt{github.com/Philip-Bachman/ICML-2015}. The code provides full details on learning rates and other hyper-parameters.

\begin{figure*}
\begin{center}
  \subfigure[]{\includegraphics[scale=0.4]{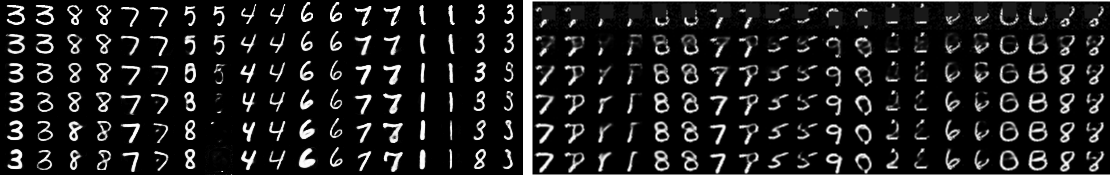}\vspace*{-5mm}}\vspace*{-3mm}
  \subfigure[]{\includegraphics[scale=0.25]{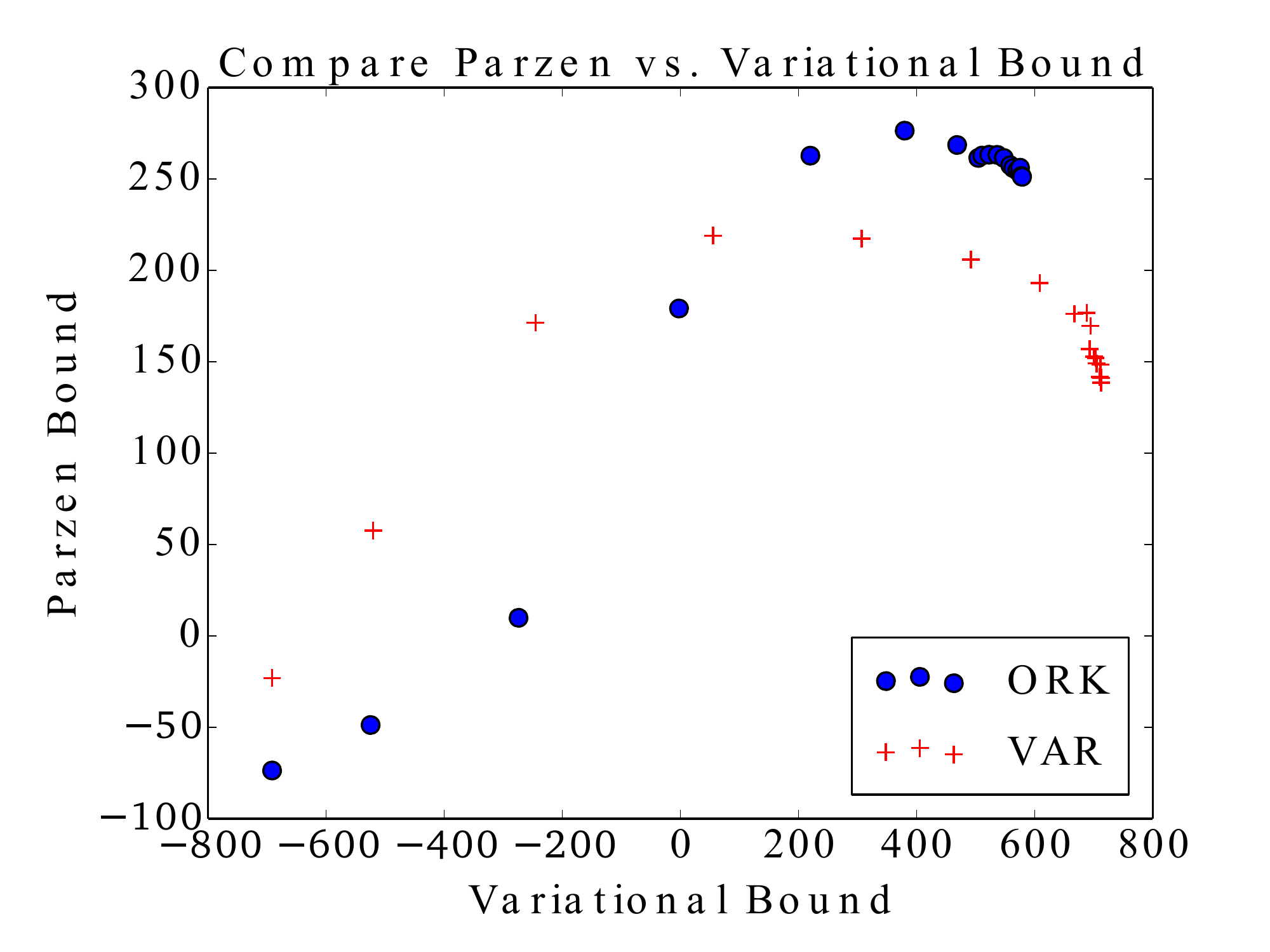}\vspace*{-3mm}}
  \subfigure[]{\includegraphics[scale=0.25]{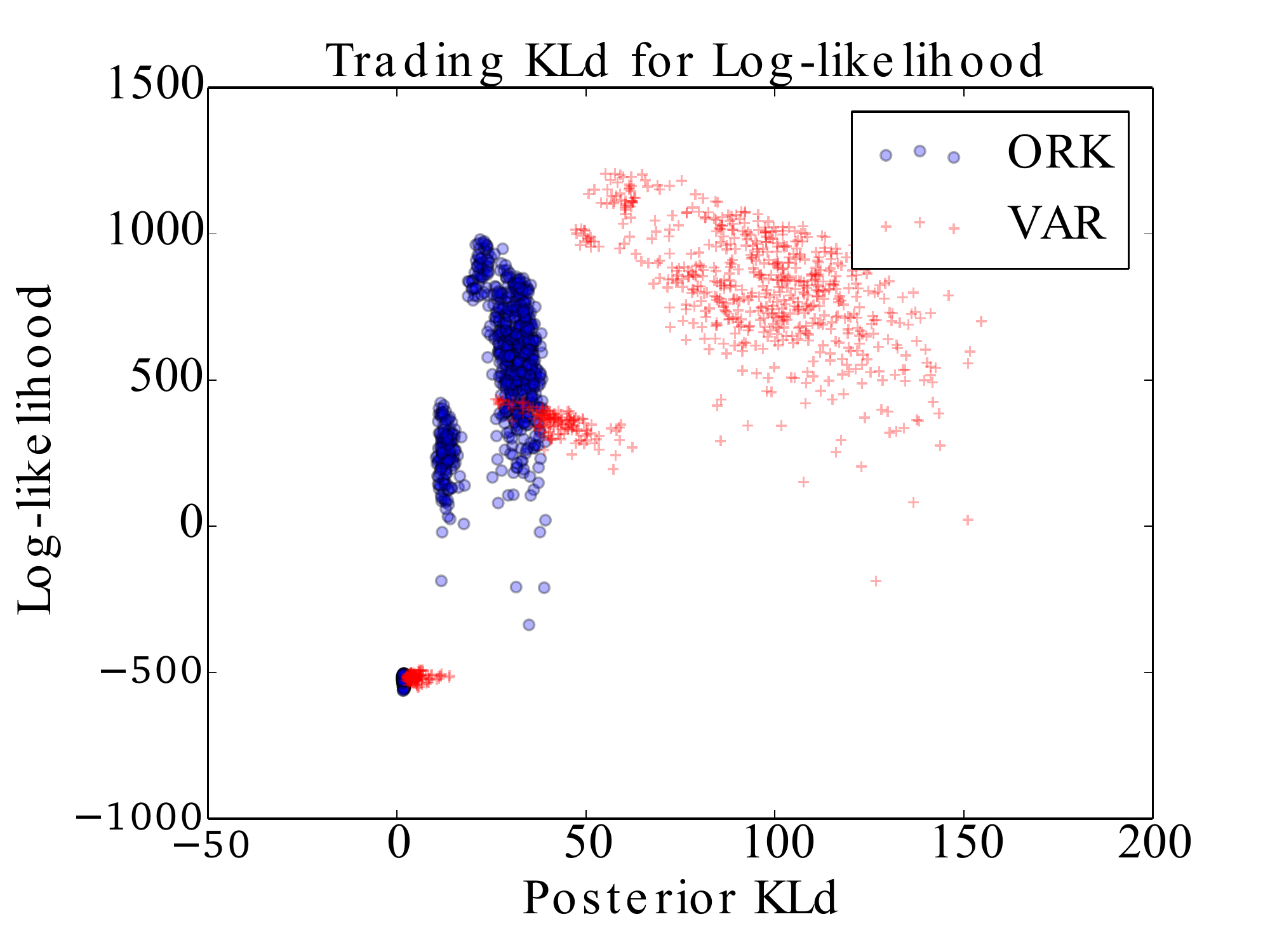}\vspace*{-3mm}}
  \subfigure[]{\includegraphics[scale=0.30]{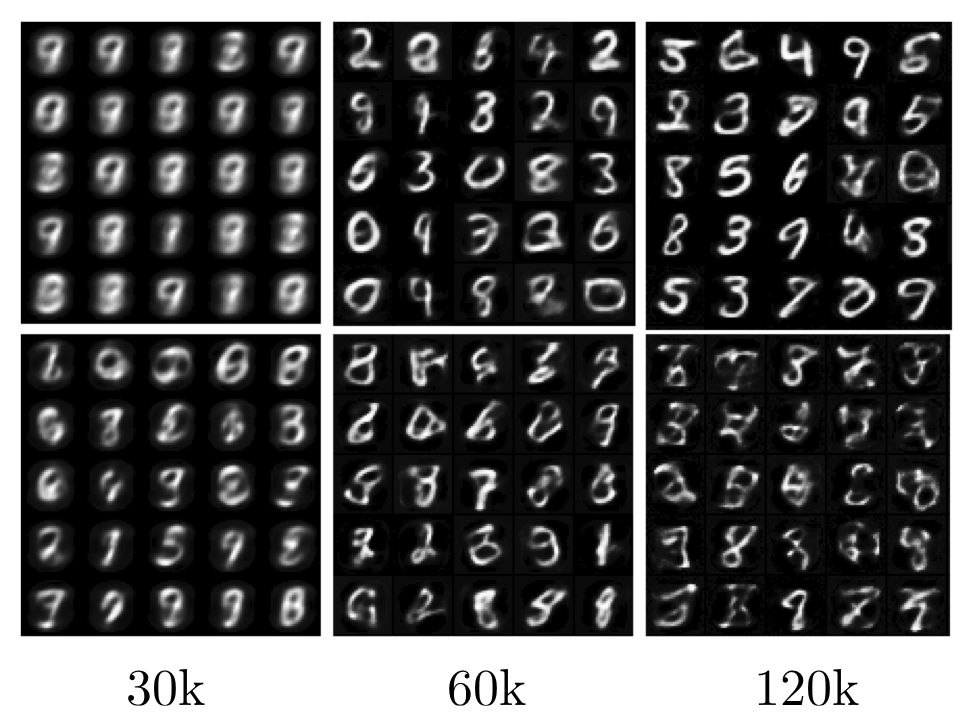}\vspace*{-3mm}}\vspace*{-3mm}
  \subfigure[]{\includegraphics[scale=0.38]{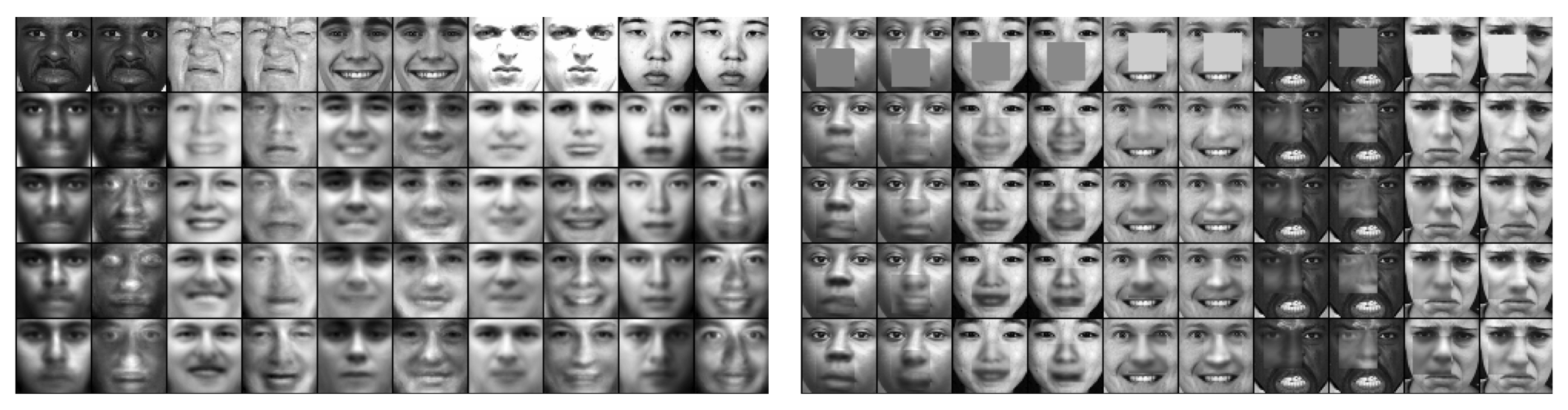}\vspace*{-3mm}}\vspace*{-3mm}
  \subfigure[]{\includegraphics[scale=0.25]{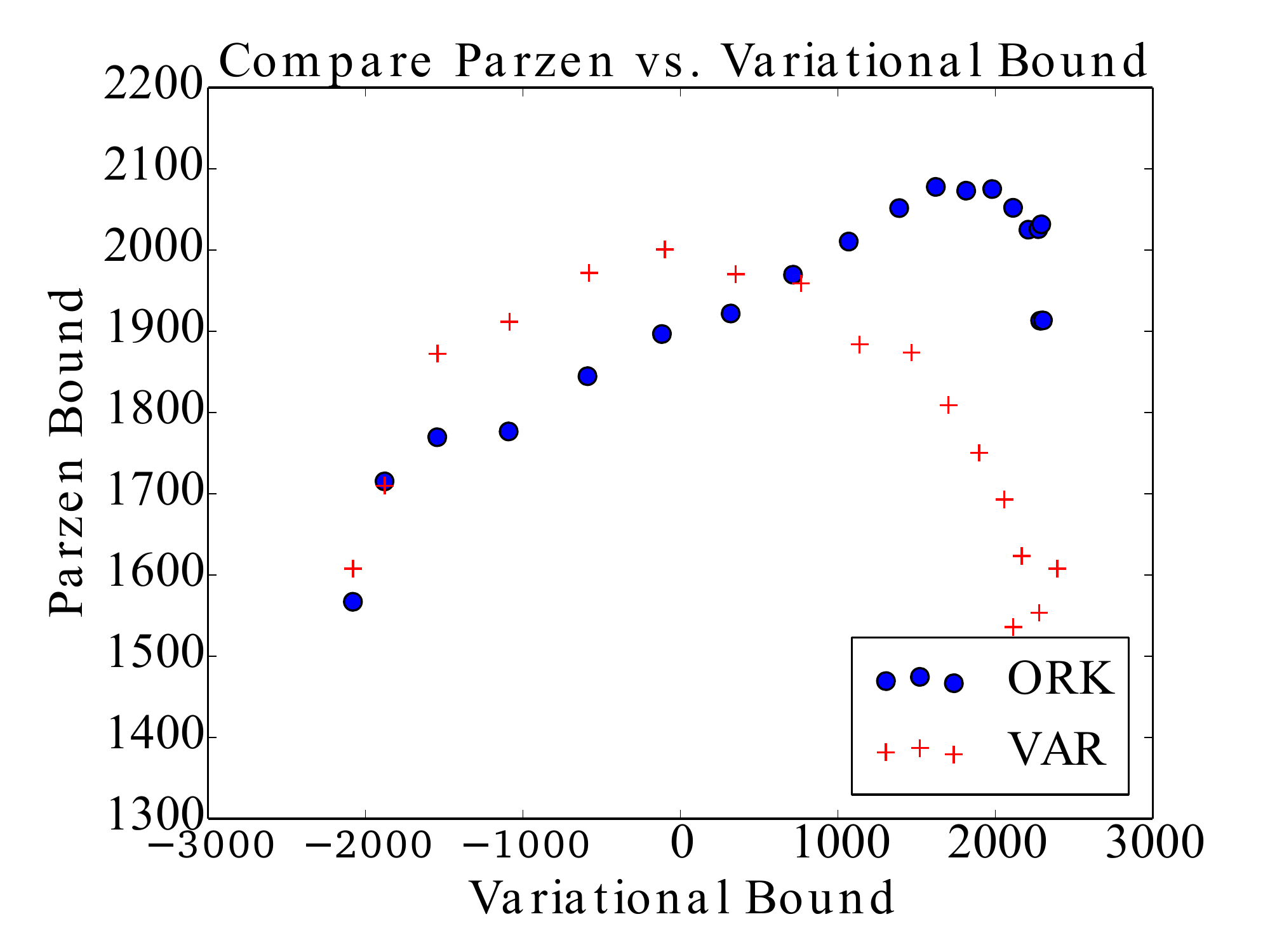}}
  \subfigure[]{\includegraphics[scale=0.25]{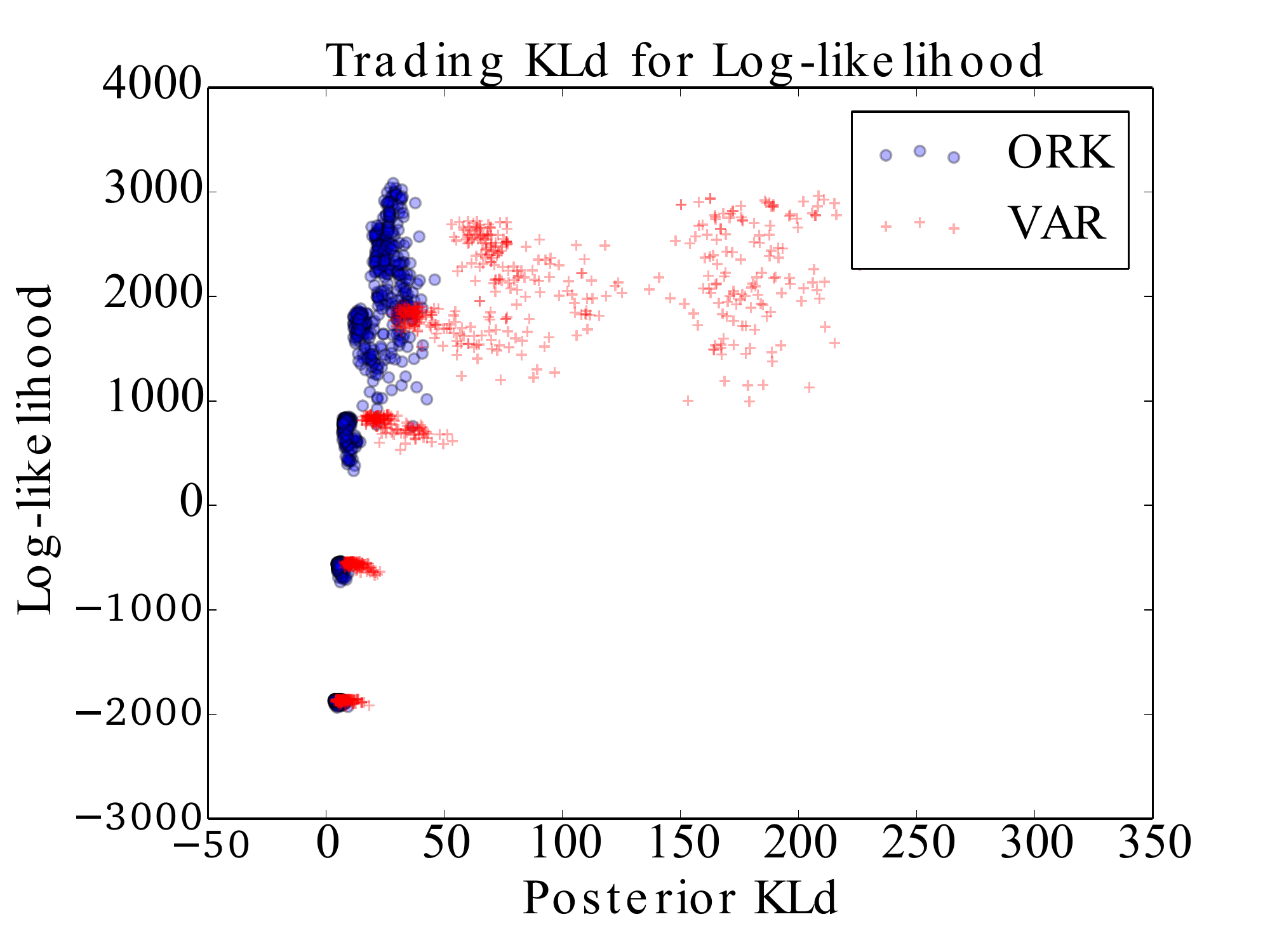}}
  \subfigure[]{\includegraphics[scale=0.26]{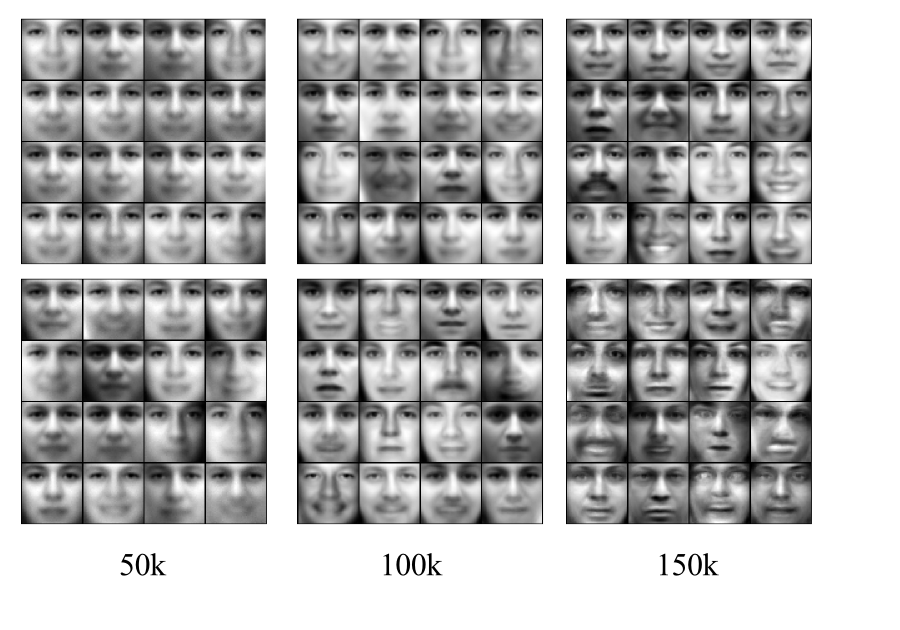}}\vspace*{-3mm}
  \caption{MNIST and TFD results. (a) compares the behavior of chains generated by models trained with over-regularized posterior KL divergence (ORK) and chains generated by models trained to minimize the standard variational free energy (VAR). The top row gives the digit image with which each chain was initialized, and the left/right chain in each pair sharing the same initialization represent chains generated by the ORK/VAR models. For the left block of examples the chains were allowed to run ``freely'' after initialization, and we show every 5th sample generated by the chains. For the right block the chains were run under ``partial control'', wherein some subset of the pixels were held fixed at their initial values while the remaining pixels were occluded at initialization and ``painted in'' by the model over multiple steps. We show every 2nd generated sample. (b) shows how the ORK and VAR models perform with respect to bounds on the validation set log-likelihood provided by the Gaussian Parzen density estimator described in \cite{Breuleux2011} and the variational free energy described in Eq.~\ref{eq:var_free_energy_1}. We computed these values after every 10k parameter updates. (c) shows the joint distribution over the reconstruction and posterior KL terms (i.e.~$\log p_{\theta}(x | z)$ and $\KL(q_{\phi}(z|x) || p_{\ast}(z))$) in Eq.~\ref{eq:var_free_energy_1} measured at several points during the training of the ORK and VAR models. We computed these values after every 30k parameter updates by averaging over 10 passes through the system comprising $q_{\phi}$ and $p_{\theta}$ for each of 150 examples selected at random from a validation set. (d) shows independent samples from the ORK (top row) and VAR (bottom row) models after 30k, 60k, and 120k parameter updates. (e)-(h) provide results for TFD analogous to those in (a)-(d) for MNIST, with differences in the tests/models detailed in the main text.}
\end{center}
\label{fig:mnist_and_tfd_results}
\end{figure*}

Fig.~3 illustrates the significantly improved long-run sampling behavior of chains trained with unrolling. The chains in Fig.~3 start at the top left and run from left-to-right and top-to-bottom. The true samples from $\DD$ used to initialize the chains are in  the top-left corners of (a)/(b). We downsampled the samples emitted by these chains 5x for this figure. Training with unrolling and collaborative guidance allows the chain to continually generate clean samples while exhibiting rapid mixing between the modes of the target distribution. Without explicit unrolling during training, the chain can only perform a few steps before degrading into samples that are well-separated from the target distribution. Qualitatively, the samples generated by the model trained with collaboratively-guided unrolling compare favourably with those presented for GSNs in~\cite{Bengio2014}.

For our second tests with the MNIST images, we used networks with roughly the same structure as in our first tests but we set $\ZZ$ to $\Real^{50}$, we used 1000 units in each of the hidden layers in $q_{\phi}/p_{\theta}$, and we used a Gaussian reconstruction distribution $p_{\theta}(x | z)$. To effect this final change, we interpreted the vector produced by the output layer of $p_{\theta}$ as the mean of a Gaussian distribution over $\XX$ and we shared a single ``bias'' parameter across all $z$ to model the element-wise log-variance of $p_{\theta}(x | z)$. We thus modeled the target distribution using an infinite mixture of isotropic Gaussians, with the mixture weights of each Gaussian fixed a priori and with their individual locations and shared scale adjusted to match the training data.

For Fig.~4(a)-(d), we trained a pair of models. We trained the first model with $\lambda = 4$ and $\gamma = 0.1$ in Eq.~\ref{eq:dispersion_reg}. We refer to this model as ORK, for over-regularized $\KL$. We trained the second model to minimize the basic free energy in Eq.~\ref{eq:var_free_energy_1}. We refer to this model as VAR. As in our first MNIST experiments, we initialized each model with pre-training by running the model in Fig.~1(c) a single step. We performed 80k pre-training updates using mini-batches and SGD updates as described for our first experiments. We then performed another 120k updates by unrolling the  chains for 6 steps following the graph in Fig.~2. The guide function $f_{\psi}$ was trained as in our first experiments.

Fig.~4 illustrates interesting behaviors of the ORK and VAR models. We found that the ORK model with strong regularization on $\KL(q_{\phi}(z | x) || p_{\ast}(z))$ was able to significantly out-perform the VAR model according to the Gaussian Parzen density estimator (GPDE) test described in \cite{Breuleux2011}\footnote{Briefly, this test approximates the distribution of a generative model by drawing 10k samples from the model and then using those samples as the mixture means for a uniformly-weighted mixture of 10k Gaussians, with a shared isotropic variance selected for the mixture components based on a validation set}. On the test set, the best ORK model scored 265, which compares favorably to the 214 in~\cite{Bengio2014} and the 225 in~\cite{Goodfellow2014}. The best VAR model scored 220. The peak performance by this metric occurred much earlier in training for the VAR model than for the ORK model. Using the same network architecture, but with $\lambda$ in Eq.~\ref{eq:dispersion_reg} increased to 24, our approach achieved a score of 330 on the GPDE test.

Interestingly, the GPDE log-likelihood bound began to decrease rapidly beyond a certain point in training. However, the variational bound continued to increase. Qualitatively, this behavior is clearly reflected in the samples shown in Fig.~4(d), which we drew directly from the models by sampling from their priors $p_{\ast}(x)$. Samples generated from the ORK model remain reasonable throughout training, but eventually suffer on the GPDE bound due to excess ``sharpness''. Samples drawn from the VAE model after 150k updates, when the VAE model significantly outperforms the ORK model in terms of the variational bound, are hardly recognizable as handwritten digits. In effect, the model is concentrating its posterior mass, as given by $q_{\phi}(z | x)$, on increasingly small regions of $\ZZ$, in exchange for significant reductions in the reconstruction cost $-\log p_{\theta}(x | z)$. This seems to lead to most of the mass of $p_{\ast}(z)$ falling on $z$s which have little or no mass under any of the $q_{\phi}(z | x)$. By forcing the $q_{\phi}(z | x)$ to be more dispersed over $\ZZ$, our added $\KL$ terms in Eq.~\ref{eq:dispersion_reg} help mitigate this issue, albeit at the cost of less precise reconstruction of any particular digit. The scatter plots in Fig.~4 illustrate the evolution of the GPDE/variational bounds over the course of training and the trade-off between reconstruction cost and posterior $\KL$ that is obtained by the ORK and VAR models.

We performed analogous tests with the TFD dataset, which comprises 48x48 grayscale images of frontal faces in various expressions. We made a few changes from the second set of MNIST tests. We expanded the hidden layers of the networks representing $q_{\phi}/p_{\theta}$ to 2000 rectified-linear units each and we expanded the latent space $\ZZ$ to $\Real^{100}$. We preprocessed the images to have pixel intensities in the range $[0...1]$, as in~\cite{Bengio2014, Goodfellow2014}. We extended the pre-training phase to 150k updates and the unrolled, collaboratively-guided phase was reduced to 60k updates. Interestingly, in these tests the ORK model did not suffer significantly in terms of the variational bound, while achieving dramatically improved performance on the GPDE bound. For comparison, best previous results on the GPDE bound for this dataset are 2050 \cite{Goodfellow2014} and 2110 \cite{Bengio2013}. Our ORK model scored 2060 on the test set using a GPDE variance selected on the validation set. When we increased $\lambda$ for the ORK model from 5 to 30, the GPDE score increased to 2130.

\section{Discussion}
\label{sec:discussion}

We presented an approach for learning generative models belonging to a simple subset of GSNs, using variational auto-encoders as building blocks. We generated Markov chains by looping the output of these auto encoders back into the input, and trained them to generate random walks along a target manifold, based on feedback from a guide function trained to discriminate between samples emitted by the chain and samples drawn from the data manifold. 
A key conceptual contribution of our approach is that we run the generative process as an unrolled Markov chain according to its a natural dynamics, i.e. the same way we want to run it ``in the wild", and then correct differences between exhibited and desired behavior by providing direct feedback about their magnitude and location (instead of trying to force the behavior in some way during the generation process). The experimental evaluation demonstrates that this direct approach is beneficial in practice.

In the long run, we believe it will be interesting to focus on interpreting our method from a reinforcement learning point of view. In this case, the ``ease of modeling" the posterior distribution $p(x|z)$ can be viewed as a reward to be maximized, and may be easily replaced or augmented with other sources of reward. The current approach of using back propagation through time for the training could also be replaced by more efficient methods based on eligibility traces. 
While we only considered generating random walks over manifolds in this paper, in future work we would like to apply our approach to modeling distributions over observed trajectories on manifolds, e.g., as seen in speech, video, motion capture, and other sequential data.

\subsection*{Acknowledgements}

Funding for this work was provided by NSERC. The authors would also like to thank the anonymous reviewers for providing helpful feedback.

\bibliography{icml_2015_paper}
\bibliographystyle{icml2015}

\clearpage


\end{document}


\maketitle

\subsection{Explaining the Variational Free-Energy}


Given distributions $p_{\theta}(x | z)$, $q_{\phi}(z | x)$, and $p_{\ast}(z)$, we can define several \emph{derived distributions}:
\begin{eqnarray}
p_{\theta}(x; p_{\ast})\!\! \!\!\!&=& \!\! \!\!\!\sum_{z} p_{\theta}(x | z) p_{\ast}(z)\\
p_{\theta}(z | x; p_{\ast})\!\! \!\!\! &=&\!\! \!\!\! \frac{p_{\theta}(x | z) p_{\ast}(z)}{p_{\theta}(x; p_{\ast})}\\
p_{\theta}(x, z; p_{\ast}) \!\! \!\!\!&=&\!\! \!\!\! p_{\theta}(x | z) p_{\ast}(z) = p_{\theta}(z | x; p_{\ast}) p_{\theta}(x; p_{\ast})
\end{eqnarray}
 Given these distributions, we now work ``backwards'' from $\log p_{\theta}(x; p_{\ast})$:
\begin{eqnarray}
\, &\log& p_{\theta}(x; p_{\ast}) = \sum_{z} q_{\phi}(z | x) \log p_{\theta}(x; p_{\ast}) \label{eq:vfe_1} \\
\, &=& \sum_{z} q_{\phi}(z | x) \log \frac{p_{\theta}(z | x; p_{\ast}) p_{\theta}(x; p_{\ast})}{p_{\theta}(z | x; p_{\ast})} \label{eq:vfe_2} \\
\, &=& \sum_{z} q_{\phi}(z | x) \log \frac{p_{\theta}(x, z; p_{\ast})}{p_{\theta}(z | x; p_{\ast})} \label{eq:vfe_3} \\
\, &=& \sum_{z} q_{\phi}(z | x) ( \log p_{\theta}(x, z; p_{\ast}) - \log q_{\phi}(z | x) \label{eq:vfe_4} \\ 
\, &\;& \tenspaces + \log q_{\phi}(z | x) - \log p_{\theta}(z | x; p_{\ast}) ) \nonumber \\
\, &=& \sum_{z} q_{\phi}(z | x) \log \frac{p_{\theta}(x, z; p_{\ast})}{q_{\phi}(z | x)} + \label{eq:vfe_5} \\
\, &\;& \tenspaces \sum_{z} q_{\phi}(z | x) \log \frac{q_{\phi}(z | x)}{p_{\theta}(z | x; p_{\ast})} \nonumber \\
\, &=& \sum_{z} q_{\phi}(z | x) \left( \log p_{\theta}(x | z) + \log \frac{p_{\ast}(z)}{q_{\phi}(z | x)} \right) + \label{eq:vfe_6} \\
\, &\;& \tenspaces \KL(q_{\phi}(z | x) || p_{\theta}(z | x; p_{\ast})) \nonumber \\
\, &\geq& \sum_{z} q_{\phi}(z | x) \log p_{\theta}(x | z) - \KL(q_{\phi}(z | x) || p_{\ast}(z)) \fivespaces \label{eq:vfe_7} \\
\, &\geq& -\FF(x; q_{\phi}, p_{\theta}, p_{\ast}) \label{eq:vfe_8}
\end{eqnarray}
where Eqns.~\ref{eq:vfe_7}-\ref{eq:vfe_8} define the variational free-energy:
\begin{eqnarray}
\FF(x; q_{\phi}, p_{\theta}, p_{\ast}) &=& -\sum_{z} q_{\phi}(z | x) \log p_{\theta}(x | z) \label{eq:vfe_9} \\
\, &\,& \fivespaces \;\; + \KL(q_{\phi}(z | x) || p_{\ast}(z)) \nonumber \\
\, &\geq& -\log p_{\theta}(x; p_{\ast})
\end{eqnarray}
These equations follow from simple algebraic manipulation and Eqn.~\ref{eq:vfe_6}-\ref{eq:vfe_7} comes from non-negativity and definition of $\KL$.
In this derivation of the variational free-energy, we treated $p_{\theta}$, $q_{\phi}$, and $p_{\ast}$ simply as computational mechanisms for producing valid distributions over the appropriate spaces. This emphasizes the fact that, for any triplet of distributions $(p_{\theta}, q_{\phi}, p_{\ast})$, $\FF(x; q_{\phi}, p_{\theta}, p_{\ast})$ can be computed and gives a lower-bound on $\log p_{\theta}(x; p_{\ast})$ for the \emph{derived distribution} $p_{\theta}(x; p_{\ast})$.

Note that the derived distributions we used result strictly from interactions between $p_{\theta}(x | z)$ and $p_{\ast}(z)$, and are independent of $q_{\phi}(z | x)$. Therefore, we could change the domain of $q_{\phi}(z | x)$ to some alternate, arbitrary space $\YY$ such that $q_{\phi}(z | y)$ produces distributions over $\ZZ$ given inputs from $\YY$. Plugging such a $q_{\phi}$ into Eqn.~\ref{eq:vfe_9} (and substituting some $y$s for some $x$s appropriately) still produces a valid free-energy $\FF(x, y; q_{\phi}, p_{\theta}, p_{\ast})$, which still upper-bounds the negative log-likelihood of $x$ under  $p_{\theta}(x; p_{\ast})$. We refer to the resulting system comprising $q_{\phi}(z | y)$, $p_{\theta}(x | z)$, and $p_{\ast}(z)$ as a variational transcoder.

Variational transcoding is a very general mechanism which encompasses standard variational auto-encoders, where $\YY = \XX$, and methods for sequence-to-sequence learning, image-to-text generation, Bayesian classification, etc. E.g., the sequence-to-sequence learning method in \cite{Sutskever2014} can be interpreted as a variational transcoder in which $p_{\theta}(x | z)/q_{\theta}(z | y)$ are constructed from LSTMs, $p_{\ast}(z)$ is, e.g., an isotropic Gaussian distribution over $\ZZ$, and the distributions output by $q_{\phi}(z | y)$ are fixed to be Dirac deltas over $\ZZ$. 